\documentclass[lettersize,journal]{IEEEtran}
\usepackage{amsmath,amsfonts}
\usepackage{algorithmic}
\usepackage{algorithm}
\usepackage{array}
\usepackage[caption=false,font=normalsize,labelfont=sf,textfont=sf]{subfig}
\usepackage{textcomp}
\usepackage{stfloats}
\usepackage{url}
\usepackage{verbatim}
\usepackage{graphicx}
\usepackage{cite}
\usepackage{amsthm}
\usepackage{multirow}
\newtheorem{definition}{Definition}
\begin{document}

\title{SVFit: Parameter-Efficient Fine-Tuning of Large Pre-Trained Models Using Singular Values}

	\author{Chengwei Sun, Jiwei Wei$^{\ast}$ \thanks{*Corresponding author}, Yujia Wu, Yiming Shi, Shiyuan He, Zeyu Ma, Ning Xie, and Yang Yang
		 \thanks{Chengwei Sun, Jiwei Wei, Yujia Wu, Yiming Shi, Shiyuan He, Zeyu Ma, Ning Xie, and Yang Yang are with the Center for Future Media and School of Computer Science and Engineering, University  of Electronic Science and Technology of China, Chengdu 611731, China (e-mail: suncw10@126.com; mathematic6@gmail.com; 202322080314@std.uestc.edu.cn; yimingshi666@gmail.com; shiyuanhe.david@gmail.com; cnzeyuma@163.com; seanxiening@gmail.com; yang.yang@uestc.edu.cn).}
	}



\maketitle

\begin{abstract}
Large pre-trained models (LPMs) have demonstrated exceptional performance in diverse natural language processing and computer vision tasks. However, fully fine-tuning these models poses substantial memory challenges, particularly in resource-constrained environments. Parameter-efficient fine-tuning (PEFT) methods, such as LoRA, mitigate this issue by adjusting only a small subset of parameters. Nevertheless, these methods typically employ random initialization for low-rank matrices, which can lead to inefficiencies in gradient descent and diminished generalizability due to suboptimal starting points. To address these limitations, we propose SVFit, a novel PEFT approach that leverages singular value decomposition (SVD) to initialize low-rank matrices using critical singular values as trainable parameters.  Specifically, SVFit performs SVD on the pre-trained weight matrix to obtain the best rank-\(r\) approximation matrix, emphasizing the most critical singular values that capture over 99\% of the matrix's information. These top-\(r\) singular values are then used as trainable parameters to scale the fundamental subspaces of the matrix, facilitating rapid domain adaptation. Extensive experiments across various pre-trained models in natural language understanding, text-to-image generation, and image classification tasks reveal that SVFit outperforms LoRA while requiring 16 times fewer trainable parameters.
\end{abstract}


\section{Introduction}
\IEEEPARstart{L}{arge} pre-trained models (LPMs), such as RoBERTa~\cite{liu2019roberta} with an impressive 125 million trainable parameters, ViT~\cite{vit} boasting 354 million parameters, and LLaMA~\cite{touvron2023llama} featuring 700 million to 65 billion parameters, have become indispensable tools in natural language processing~\cite{LLM1,LLM2,LLM3} and computer vision~\cite{82,83,84,T5,T2,T3,T6}, showcasing remarkable performance across a spectrum of tasks. The industry's relentless pursuit of scaling up model parameters to the billion or even trillion range continues to push the boundaries of large models~\cite{LLM8,llm9,hambardzumyan2021warp}. Nonetheless, the immense size and computational demands of these models present significant challenges for adapting them to specific downstream tasks, particularly in resource-constrained environments~\cite{wolf2020transformers,LLM10}.

In response to this challenge, parameter-efficient fine-tuning (PEFT) methods have emerged as a promising solution to reduce memory requirements~\cite{lora,zi2023delta,vera,Prompttuning2,MELORA,Dylora}. These methods focus on updating a limited subset of parameters—either a portion of the existing model parameters or an entirely new set~\cite{peft3,Prefix-tuning,Prompttuning1}. The main goal is to retain the knowledge embedded in LPMs while adapting them to specific tasks, thereby minimizing the risk of catastrophic forgetting. Among these methods, Low-Rank Adaptation (LoRA)~\cite{lora} has garnered particular attention for its stable performance across diverse downstream tasks. LoRA is based on the hypothesis that LPMs can still learn effectively when projected onto a smaller subspace due to the low intrinsic dimensionality of the tasks~\cite{lowrank1}. It introduces two low-rank matrices, \(A\) and \(B\), to approximate weight updates during fine-tuning. Specifically, LoRA initializes \(A\) using a Gaussian distribution and sets \(B\) to zero, as illustrated in Fig. \ref{fig:1}(a). This initialization scheme enables incremental updates to be seamlessly integrated into the pre-trained weights without causing significant delays in inference. Building on LoRA's foundation, several subsequent approaches~\cite{Lora-fa,zi2023delta,vera,MELORA} have adhered to this paradigm, experimenting with various initialization strategies such as Kaiming uniform~\cite{RAN1} , and further enhance computational efficiency. However, this raises the question of whether random initialization is indeed optimal. Specifically, random initialization may prevent the main components of the pre-trained weight matrix \(W\) from being effectively updated during the initial stages of fine-tuning. This limitation could lead to inefficiencies in gradient descent, potentially resulting in suboptimal local minima and impairing the generalization performance of the model.


 \begin{figure*}[t]
	\centering
    \includegraphics[width=7in]{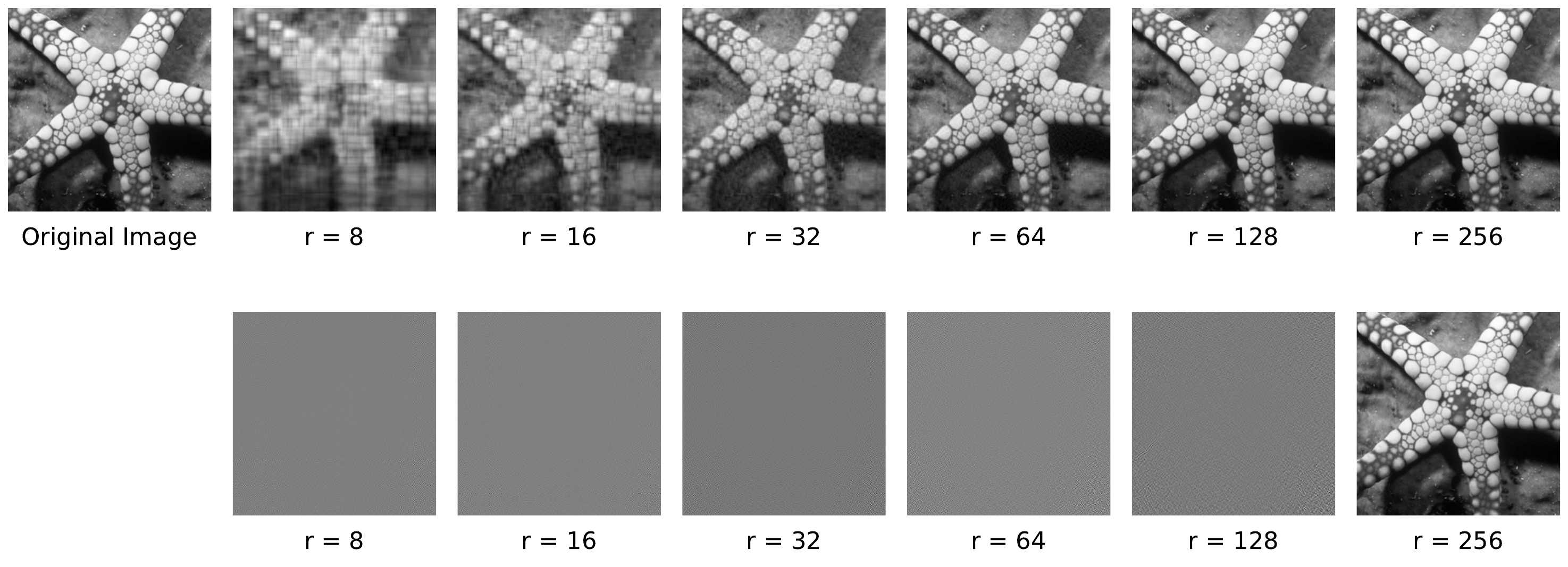}
	\caption{SVD-based reconstruction results of the Fishstar image ($256 \times 256$)~\cite{DATA}. The first row shows the reconstruction using the top 8, 16, 32, 64, 128, and 256 largest singular values, sorted in descending order (\(r = 8\), \(r = 16\), \(r = 32\), \(r = 64\), \(r = 128\), \(r = 256\)). The second row displays the reconstruction using the smallest 8, 16, 32, 64, 128, and 256 singular values, sorted in ascending order. This comparison highlights the pivotal role of dominant singular values in maintaining image quality, while the smallest singular values have minimal impact on the overall structure.}
 \label{fig:0}
\end{figure*}

 This paper introduces SVFit, an innovative PEFT strategy for LPMs. SVFit utilizes a distinctive initialization strategy by applying SVD to the pre-trained weight matrix \(W\), resulting in two components: the best rank-\(r\) approximation matrix \(W_{r}\) and a residual matrix \(W_{e}\), with \(W_{e}\) capturing the smaller singular values. Our analysis verifies that the top 10\%, or even 1\%, of singular values contribute to over 99\% of the total matrix sum. As depicted in Fig. \ref{fig:0}, SVD was performed on the Fishstar, with the first set of experiments arranging the singular values in descending order and reconstructing the images using the largest 8, 16, 32, 64, 128, and 256 singular values, as shown in the first row. The second set of experiments arranged the singular values in ascending order, reconstructing the images with the smallest 8, 16, 32, 64, 128, and 256 singular values, with results displayed in the second row. These findings underscore the critical role of larger singular values in preserving image quality. Therefore, the pre-trained weight matrix \(W\) can be effectively approximated using only the most significant singular values above a threshold \(r\), while the smaller singular values contribute minimally to the overall structure. As a result, \(W_{r}\) retains the essential pre-trained knowledge, and \(W_{e}\) is kept frozen during training. Additionally, inspired by recent advances in image generation that demonstrate the effectiveness of learnable scaling factors for improved domain adaptation \cite{xie2023difffit}, SVFit uses the most critical singular values obtained from SVD as trainable parameters. Only the top \(r\) singular values \(\Sigma_{r}\) within \(W_{r}\) are trained, with the fundamental subspaces derived from SVD scaled to promote rapid adaptation to new domains, as illustrated in Fig. \ref{fig:1}(c) and Fig. \ref{fig:subspace}. This approach enhances the learning of new domain knowledge for downstream tasks while preserving pre-trained information and significantly reducing the number of trainable parameters.

Extensive experiments have been conducted to verify the effectiveness of SVFit across various tasks and models. Specifically, RoBERTa-base and RoBERTa-large were used for natural language understanding tasks, ViT-base and ViT-large were used for image classification tasks, and  Stable Diffusion v1.5 was used for subject-driven text-to-image task. The experimental results demonstrate that SVFit achieves superior performance with a significantly reduced number of parameters. For instance, in the image classification task using the ViT-large model, SVFit outperforms LoRA with only 0.036M trainable parameters, compared to LoRA's 0.8M.

The primary contributions of this paper are as follows:

\begin{itemize}
\item We present SVFit, a novel PEFT method that enhances the initialization process by utilizing SVD to initialize low-rank matrices derived from pre-trained weights. SVFit focuses on training only the most significant top-$r$ singular values, significantly reducing the number of trainable parameters while achieving efficient fine-tuning and preserving the model's core capabilities.

\item We offer a theoretical analysis to uncover the mechanisms underlying SVFit. This analysis demonstrates how leveraging singular values enables rapid adaptation by effectively capturing the essential information from pre-trained models and efficiently learning new domain-specific knowledge with minimal parameters.

\item SVFit is evaluated on a range of tasks, such as natural language understanding, image classification, and subject-driven text-to-image generation. It consistently outperforms LoRA and other recent state-of-the-art techniques in terms of parameter efficiency and overall performance.
\end{itemize}

\section{Related Work}
\subsection{Parameter-Efficient Fine-Tuning}
With the development of LPMs, adapting models with billions of parameters to specific downstream tasks has become increasingly challenging due to their complexity and computational demands~\cite{LLM4,LLM5,LLM6,llm7}. Parameter-efficient fine-tuning (PEFT) has garnered significant attention in recent years for its ability to minimize the parameters and memory requirements needed while maintaining efficiency and accuracy, achieving performance comparable to full fine-tuning. Certain PEFT methods achieve fine-tuning by incorporating supplementary modules or optimizing prompts and prefixes. For instance, Adapter~\cite{adap} integrates lightweight trainable parameters between pre-trained layers while maintaining fixed pre-trained weights. Prefix tuning~\cite{Prefix-tuning} involves appending prefix parameters to the hidden states across all layers of the model. Prompt tuning~\cite{Prompttuning1} utilizes templates to reconstruct prompts, updating only parameters relevant to prompt comprehension. Despite their significant performance gains, these approaches unavoidably introduce additional overhead during inference.

\subsection{Low-Rank Adaptation}
\begin{figure*}[t]
	\centering
   \includegraphics[width=7in]{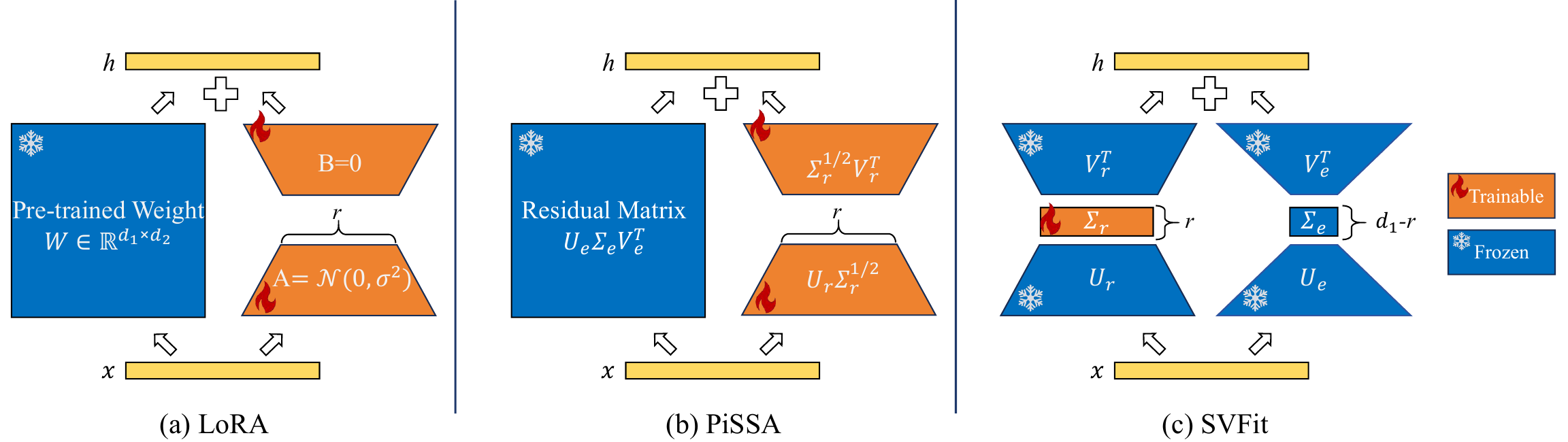}
	\caption{Visual comparison among LoRA,  PiSSA, and SVFit. (a) LoRA introduces two low-rank matrices $A$ and $B$ to approximate weight updates during fine-tuning. (b) PiSSA initializes $A$ and $B$ with the principal components of the pre-trained weight $W$, freezing the residual matrix during fine-tuning. (c) SVFit initializes low-rank matrices through SVD of $W$ and trains only the most significant top-$r$ singular values (for simplicity, $d_{1} \ll  d_{2}$ is assumed). }
		\label{fig:1}
\end{figure*}

Low-Rank Adaptation (LoRA)~\cite{lora} introduces two low-rank matrices to approximate weight updates during fine-tuning, seamlessly integrating incremental updates into pre-trained weights without causing noticeable delays in inference. 
To be specific, when provided with a pre-trained weight matrix $W \in \mathbb{R} ^{d_{1} \times d_{2}} $,  after full fine-tuning on a specific domain task, the new weight matrix is $W+W^{'}$, where $W^{'} $ represents the update containing domain knowledge. LoRA is designed to progressively update $W^{'} $ and break down $W^{'} $ through the matrix multiplication of two low-rank matrices $A$ and $B$,
\begin{equation}
    W^{'} =AB,
\end{equation}
where $A \in \mathbb{R} ^{d_{1}\times r} $, and $B \in \mathbb{R} ^{r\times d_{2}} $ with intrinsic rank $r \ll \min ( d_{1},d_{2}) $.  For $h=Wx$, the modified computation of the forward pass can be represented as 
\begin{equation}
    h=\left ( W+W^{'} \right ) x=\left ( W+AB \right )x .
\end{equation}
 In the initial training phase, $A$ undergoes random Gaussian initialization, while $B$ is initialized to zeros, as shown in Fig. \ref{fig:1}(a). This method freezes $W$ and specifically updates $A$ and $B$, significantly reducing the number of trainable parameters for downstream tasks compared to full fine-tuning. Additionally, LoRA integrates the values of matrices $A$ and $B$ into $W$ during the inference phase, ensuring that this adaptation does not introduce additional delays.


\subsection{LoRA's Variants}
Building upon LoRA, several studies have proposed modifying the parameter update strategy. LoRA-FA~\cite{Lora-fa} freezes the low-rank matrix \( A \) within LoRA, significantly reducing trainable parameters and activation memory costs without increasing computational overhead. Delta-LoRA~\cite{zi2023delta} updates both low-rank matrices \( A \) and \( B \), propagating these changes to the pre-trained weight \( W \) using the delta of their product. PiSSA~\cite{meng2024pissa} initializes  \( A \) and \( B \) with the principal components of the original matrix $W$ and places the remaining components into a residual matrix, which is kept frozen during fine-tuning. Another strategy to improve LoRA is to allow for the adaptable adjustment of LoRA rank. AdaLoRA~\cite{zhang2023adalora} utilizes SVD to parameterize incremental updates and dynamically distributes the parameter budget among weight matrices based on their importance score. SoRA~\cite{sora} employs an optimizable gate with a proximal gradient method to control sparsity, expanding the optimization space and improving parameter efficiency. Similarly, Zhang et al.~\cite{zhang2023increlora} suggest IncreLoRA, an incremental parameter allocation method that adaptively incorporates trainable parameters during training according to the importance scores of each module.


 \begin{figure*}[t]
	\centering
   \includegraphics[width=6in]{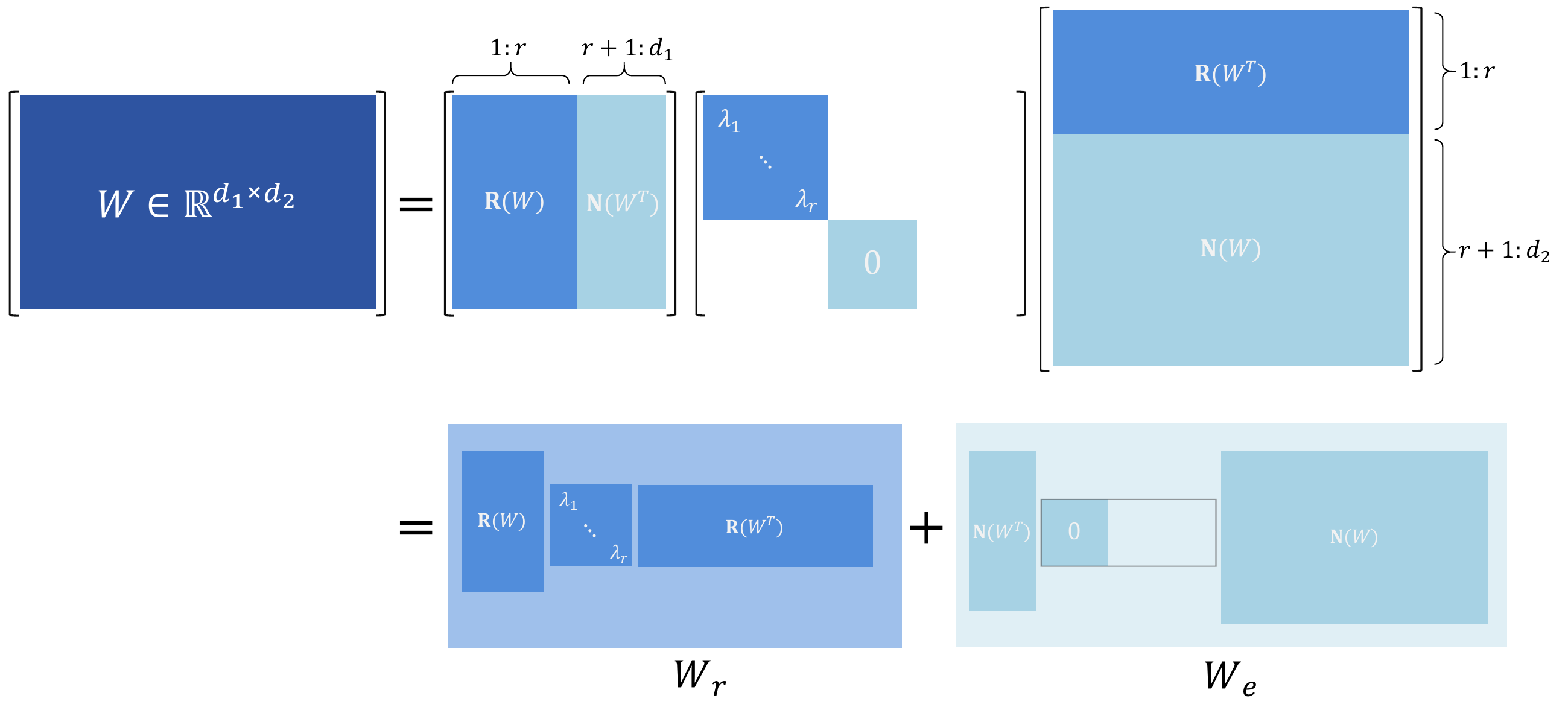}
	\caption{Illustration of the SVD of matrix \( W \) and its fundamental subspaces: This figure illustrates the SVD of the pre-trained weight matrix \( W \in \mathbb{R}^{d_{1} \times d_{2}} \), where \( W \) is decomposed into singular values and vectors as \( W = U \text{diag}(\Sigma) V^{T} \). The decomposition yields a rank-\( r \) approximation matrix \( W_{r} \) and a residual matrix \( W_{e} \). Specifically, the range space of \( W \) is spanned by \( U_{r} \), and its null space is spanned by \( V_{e} \). Conversely, the range space of \( W^{T} \) is spanned by \( V_{r} \), and its null space is spanned by \( U_{e} \).}
		\label{fig:subspace}
\end{figure*}

\section{Method}
In this section, we present SVFit, a novel PEFT approach for LPMs that leverages singular values for improved model adaptation. Unlike conventional methods that preserve the knowledge of LPMs by freezing the pre-trained weight matrix \(W\), SVFit aims to embed this knowledge into low-rank matrices and retain it permanently.  SVFit achieves this by performing SVD on  the pre-trained weight matrix  \(W\), using the most critical singular values as trainable parameters, thereby optimizing the initialization of low-rank matrices for more effective fine-tuning.

\subsection{Fundamental Subspaces Derived from SVD}\label{sec:fundamental_subspaces}
We begin by introducing key concepts related to Singular Value Decomposition (SVD) and fundamental subspaces, a technique widely used in pattern recognition and data compression~\cite{sun2023nf,9777947}. SVD transforms a dataset from a high-dimensional space to a lower-dimensional space by ranking the singular values according to their significance~\cite{svd0,svd1}. Dimensionality reduction is achieved by discarding the less significant singular values, while the remaining singular values, when combined with their corresponding singular vectors, define the reduced-dimensional space.

\begin{definition}[Range space]
Given a matrix $W\in \mathbb{R } ^{d_{1}\times d_{2}}$, the range space of matrix $W$ is the vector space spanned by the columns of $W$.  In other words, the range space is the set of all possible linear combinations of the column vectors of $W$. The range space is often denoted as $\mathbf{R} \left ( W \right ) $. Formally, the range space is defined as:
\begin{equation}
    \mathbf{R} \left ( W \right ) =\left \{ Wx\mid x\in \mathbb{R} ^{d_{2}}  \right \} \subseteq \mathbb{R} ^{d_{1}}.
\end{equation}
Similarly, the range space of $W^{T}$  is a subspace of $\mathbb{R} ^{d_{2}}$,  denoted as $\mathbf{R} \left (W^{T} \right ) $. That is,
\begin{equation}
    \mathbf{R} \left ( W^{T}  \right ) =\left \{ W^{T} y\mid y\in \mathbb{R} ^{d_{1}}  \right \} \subseteq \mathbb{R} ^{d_{2}}.
\end{equation}
\end{definition}

\begin{definition}[Null space]
Given a matrix $W\in \mathbb{R} ^{d_{1}\times d_{2}}$, the null space of matrix $W$ is the set of all vectors $x\in \mathbb{R} ^{d_{2}}$ that are mapped to the zero vector in $\mathbb{R} ^{d_{1}}$ when multiplied by $W$. Formally, the null space is defined as:
\begin{equation}
    \mathbf{N} \left ( W \right ) =\left \{ x\mid Wx=0 \right \} \subseteq \mathbb{R} ^{d_{2} }.
\end{equation}
Similarly, the null space of  matrix \( W^{T} \) consists of all vectors \( y \in \mathbb{R}^{d_{1}} \) that are mapped to the zero vector in \( \mathbb{R}^{d_{2}} \) when multiplied by \( W^{T} \). The formal definition of the null space of \( W^{T} \) is as follows:
\begin{equation}
    \mathbf{N} \left ( W^{T} \right ) =\left \{ y\mid W^{T}y=0 \right \} \subseteq \mathbb{R} ^{d_{1} }.
\end{equation}
\end{definition}

Given a matrix $W\in \mathbb{R} ^{d_{1}\times d_{2}}$ of rank $r$, its SVD is denoted as  $W=U \text{diag}( \Sigma) V^{T}$, where \( \text{diag}(\Sigma) \) is a \( d_{1} \times d_{2} \) diagonal matrix containing the singular values \( \{ \lambda_{i} \}_{1 \leq i \leq r} \) of \( W \) in descending order, with \( r \ll \min(d_{1}, d_{2}) \). The matrices \( U = [u_{1}, u_{2}, \ldots, u_{d_{1}}] \in \mathbb{R}^{d_{1} \times d_{1}} \) and \( V = [v_{1}, v_{2}, \ldots, v_{d_{2}}] \in \mathbb{R}^{d_{2} \times d_{2}} \) are orthogonal matrices. We can partition \( U \) and \( V \) into two parts by columns and denote these partitions as \( U_{r} \), \( U_{e} \), and \( V_{r} \), \( V_{e} \), respectively:
\begin{equation}
U_{r}=\left [ u_{1}, u_{2},\cdots , u_{r} \right ],  U_{e}=\left [ u_{r+1}, u_{r+2},\cdots , u_{d_{1} } \right ],
\end{equation}
\begin{equation}
V_{r}=\left [ v_{1}, v_{2},\cdots , v_{r} \right ],  V_{e}=\left [ v_{r+1}, v_{r+2},\cdots , v_{d_{2} } \right ].
\end{equation}

Then the four fundamental subspaces associated with matrix \( W \) can be obtained: 
\begin{itemize}
\item $U_{r} $ is the orthonormal basis for the range space of $W$, i.e., $\mathbf{R}(U_{r}) = \mathbf{R}(W)$;
\item $U_{e}$ is the orthonormal basis for the null space of $W^{T}$, i.e., $\mathbf{R}(U_{e}) = \mathbf{N}(W^{T})  $;
\item $V_{r} $ is the orthonormal basis for the range space of $W^{T}$, i.e., $\mathbf{R}(V_{r}) = \mathbf{R}(W^{T})$;
\item $V_{e}$ is the orthonormal basis for the null space of $W$, i.e., $\mathbf{R}(V_{e}) = \mathbf{N}(W)$.
\end{itemize}

To verify that \( \mathbf{R}(V_{e}) = \mathbf{N}(W) \), one would show that the subspace spanned by \( V_{e} \) consists precisely of the vectors that \( W \) maps to the zero vector. Similar arguments can be applied to verify the other subspaces.
\begin{proof}\label{pro:1}
	Assume the rank of the matrix $W\in \mathbb{R} ^{d_{1}\times d_{2}}$ is $r$ and $d_{1} \ll  d_{2}$. Thus, the singular values are are ordered as $\lambda _{1}\ge \lambda _{2}\ge \cdots\ge  \lambda _{r}> \lambda _{r+1}=\cdots =\lambda _{d_{1} }=0$. Since $W=U\text{diag}( \Sigma) V^{T}=\sum_{i=1}^{r} \lambda _{i} u_{i} v_{i}^{T} $, we can express $W$ as $Wx=\sum_{i=1}^{r} \lambda _{i} u_{i} v_{i}^{T}x $. Let $z=\sum_{i=r+1}^{d_{2} } \beta _{i}  v_{i}$, then $Wz=\left ( \sum_{i=1}^{r} \lambda _{i} u_{i} v_{i}^{T} 
 \right ) \left ( \sum_{i=r+1}^{d_{2} } \beta _{i}  v_{i} \right ) =0$. Therefore, the null space of $W$ is  $ \mathbf{N} \left ( W\right )=\left \{ x\mid Wx=0 \right \} =\mathbf{R} \left ( V_{e}  \right ) $.
\end{proof}

The connection between the SVD of the matrix \( W \) and the four fundamental subspaces can be depicted in Fig. \ref{fig:subspace}.

\subsection{SVFit: PEFT of LPMs Using Singular Values}
 SVFit performs SVD on the initial pre-trained weight matrix $W\in \mathbb{R } ^{d_{1}\times d_{2}}$ within the self-attention and multilayer perceptron layers to obtain the best rank-$r$ approximation 
 \begin{equation}
     W_{r}=U_{r} \text{diag}( \Sigma_{r})V_{r}^{T},
 \end{equation}
and the residual matrix
\begin{equation}
    W_{e}=U_{e} \text{diag}( \Sigma_{e})V_{e}^{T},
\end{equation}
where \(U_{r}=\left [ u_{1}, u_{2},\cdots , u_{r} \right ] \in \mathbb{R}^{d_{1} \times r}\) and \(V_{r}=\left [ v_{1}, v_{2},\cdots , v_{r} \right ] \in \mathbb{R}^{d_{2} \times r}\) are the matrices of singular vectors corresponding to the top \(r\) singular values, and \(U_{e}=\left [ u_{r+1}, u_{r+2},\cdots , u_{d_{1} } \right ]  \in \mathbb{R}^{d_{1} \times (d_{1}-r)}\) and \(V_{e}=\left [ v_{r+1}, v_{r+2},\cdots , v_{d_{2} } \right ] \in \mathbb{R}^{d_{2} \times (d_{2}-r)}\) are the matrices of singular vectors corresponding to the residual singular values. As discussed in Section~\ref{sec:fundamental_subspaces}, $\mathbf{R}(U_{r}) = \mathbf{R}(W)$, $\mathbf{R}(V_{r}) = \mathbf{R}(W^{T})$, $\mathbf{R}(U_{e}) = \mathbf{N}(W^{T})  $, and $\mathbf{R}(V_{e}) = \mathbf{N}(W)$. The singular values are arranged in descending order, with \( \Sigma_{r} = [\lambda_{1}, \lambda_{2}, \dots, \lambda_{r}] \) and \( \Sigma_{e} = [\lambda_{r+1}, \lambda_{r+2}, \dots, \lambda_{d_{1}}] \). Consequently, the pre-trained weight matrix \( W \) can be expressed as:
\begin{equation}
\begin{aligned}
    W = W_{r} + W_{e}&= U_{r} \text{diag}(\Sigma_{r}) V_{r}^{T} + U_{e} \text{diag}(\Sigma_{e}) V_{e}^{T}\\
    &=\sum_{i=1}^{r}\lambda _{i}u_{i}v_{i}^{T} + \sum_{i=r+1}^{d_{1} }\lambda _{i}u_{i}v_{i}^{T},
   \end{aligned}
\end{equation}
as shown in Fig. \ref{fig:1}(c) and Fig. \ref{fig:subspace}.

As observed in Fig. \ref{fig:0}, we performed SVD on the Fishstar to reconstruct images using various subsets of singular values: the top 8, 16, 32, 64, 128, and 256 singular values, as well as the smallest 8, 16, 32, 64, 128, and 256 singular values. The results highlight that larger singular values are crucial for preserving image quality. Notably, the top 10\% or even 1\% of singular values contribute to over 99\% of the total matrix sum. This observation indicates that the pre-trained weight matrix \( W \) can be effectively approximated by focusing on the most significant singular values above a given threshold \( r \), while the smaller singular values have minimal impact on the overall structure. Consequently, in SVFit, \( W_{e} \) is frozen during training, and the focus is placed on adapting \( W_{r} \). Inspired by recent advancements in image generation, such as learnable scaling factors for improved domain adaptation \cite{xie2023difffit}, SVFit utilizes the most critical singular values obtained from SVD as the trainable parameters. Specifically, it trains only the top \( r \) singular values \(\Sigma_{r}\) in \( W_{r} \), while scaling the fundamental subspace derived from SVD to facilitate rapid adaptation to new domains. In a word, the matrices \( U_{r} \), \( V_{r} \), \( U_{e} \), \( V_{e} \), and \(\Sigma_{e}\) are kept frozen, and the training focuses exclusively on the most significant top-\( r \) singular values \(\Sigma_{r}\), as demonstrated in Fig. \ref{fig:1}(c) and Fig. \ref{fig:subspace}. This method allows features to be projected onto a low-rank subspace defined by the orthogonal columns of \( U \) and \( V \), enabling efficient layer-wise adaptation with a reduced number of trainable parameters.

In comparison to other LoRA variants, SVFit uniquely employs the most critical singular values from SVD initialization as trainable parameters, leading to more effective learning of new domain knowledge for downstream tasks while preserving pre-trained information. This approach significantly reduces the number of trainable parameters without introducing additional computational overhead or latency during inference, as the module can be seamlessly integrated into the original matrix post-training.

\section{Experiment}
In this section, we conduct extensive experiments to evaluate SVFit in the contexts of natural language understanding and computer vision. We fine-tune the RoBERTa-base and RoBERTa-large models~\cite{liu2019roberta} on the GLUE benchmark~\cite{wang2018glue} and apply SVFit to fine-tune ViT-base and ViT-large models~\cite{vit} for image classification tasks~\cite{T1,T4}. Additionally, we fine-tune Stable Diffusion v1.5~\cite{SD} to generate diverse images of a subject instance in various environments~\cite{TEXT,ruiz2023dreambooth}. We also vary the rank in our method for one task to examine how performance scales with the number of trainable parameters and analyze the influence of the learning rate. Our experiments are conducted using PyTorch, with pre-trained weights and configuration files obtained from HuggingFace~\cite{wolf2019huggingface}, on NVIDIA A6000 GPUs.

\subsection{Baselines}
We compare SVFit with full fine-tuning and popular PEFT methods, including LoRA, DyLoRA, AdaLoRA, and PiSSA.
\begin{itemize}
\item Full fine-tuning involves updating the entire set of model parameters, initialized with pre-trained weights and biases, through gradient descent~\cite{full}. Although this approach is straightforward and robust, it demands substantial computational resources.
\item LoRA~\cite{lora} employs two low-rank matrices to learn incremental updates, reducing GPU memory cost. We replicate their experimental setup for a fair comparison.
\item DyLoRA~\cite{Dylora} dynamically selects a random rank $r$ for LoRA modules during training.
\item  AdaLoRA~\cite{zhang2023adalora} addresses the challenge of optimal rank selection for incremental updates by adaptively pruning singular values based on their magnitudes, resulting in different ranks for different layers. 
\item  PiSSA~\cite{meng2024pissa},  structurally similar to LoRA, initializes the adapter matrices \( A \) and \( B \) using the principal components of the original weight matrix \( W \), while the remaining components form a residual matrix that is kept frozen during fine-tuning.
\end{itemize}

\begin{table}[htbp]
\caption{Hyperparameter setup of SVFit for the GLUE benchmark}
\label{tab:hp1}
\centering
\setlength{\tabcolsep}{2pt} 
\begin{tabular}{ll|cccccc}
\hline
~& Hyperparameter&COLA & STS-B &RTE &MRPC &SST-2 &QNLI \\
\hline
\multirow{8}{*}{\rotatebox{90}{Base}} & Rank & \multicolumn{6}{c}{768} \\
 & Warmup Ratio & \multicolumn{6}{c}{0.06} \\
 & LR Scheduler & \multicolumn{6}{c}{Linear} \\
 & Weight Decay & \multicolumn{6}{c}{0.1} \\
 & Trainable Matrices & \multicolumn{6}{c}{$W_{Q},W_{V}$} \\
 & Max Sequence Length & \multicolumn{6}{c}{512} \\
 & Epochs & 80 & 40 & 80 & 30 & 60 & 25 \\
 & Batch Size & 64 & 64 & 16 & 64 & 64 & 64 \\
 & Learning Rate & 0.01 & 0.01 & 0.01 & 0.02 & 0.003 & 0.003  \\
\hline
\multirow{8}{*}{\rotatebox{90}{Large}} & Rank & \multicolumn{6}{c}{768} \\
 & Batch Size & \multicolumn{6}{c}{32} \\
 & Warmup Ratio & \multicolumn{6}{c}{0.06} \\
 & LR Scheduler & \multicolumn{6}{c}{Linear} \\
 & Weight Decay & \multicolumn{6}{c}{0.1} \\
 & Trainable Matrices & \multicolumn{6}{c}{$W_{Q},W_{V}$} \\
 & Max Sequence Length & \multicolumn{6}{c}{128} \\
 & Epochs & 40 & 20 & 40 & 40 & 10 & 20 \\
 & Learning Rate & 0.01 & 0.01 & 0.01 & 0.003 & 0.001 & 0.002 \\
\hline
\end{tabular}
\end{table}

\begin{table*}[t] 
\caption{Performance comparison of various fine-tuning methods on the GLUE benchmark using RoBERTa-base and RoBERTa-large models. Metrics include MCC for CoLA, PCC for STS-B, and ACC for RTE, MRPC, SST-2, and QNLI. Results are reported as the median of 5 runs with different random seeds. The highest score for each dataset is highlighted in bold, with higher values indicating better performance across all metrics}
\label{tab:GLUE}
\centering
 \setlength{\tabcolsep}{13pt}
\begin{tabular}{llcccccccc}
\hline
\multirow{2}*{~}&\multirow{2}*{Method }& \#  Trainable  & CoLA & STS-B & RTE & MRPC & SST-2 & QNLI & \multirow{2}*{Avg.} \\
 &~ &Parameters  & (MCC) & (PCC) & (ACC) & (ACC) & (ACC) & (ACC) & \\
\hline
\multirow{6}{*}{\rotatebox{90}{Base}}&FT & 125M &63.6 &91.2	&78.7	&\textbf{90.2}	&94.8 &92.8	&\textbf{85.2} \\
&LoRA & 0.3M& 63.4&	91.5&	78.4&	89.7&	\textbf{95.1}&	\textbf{93.3}	&\textbf{85.2} \\
&DyLoRA& 0.3M & 61.1&	91.1&	78.7&	89.5&	94.3&	92.2&	84.5 \\
&AdaLoRA& 0.3M  & 62&	90.5&	\textbf{81}&	88.7&	94.5&	93.1&	85.0 \\
&PiSSA& 0.3M  & 63.8&	90.8&	75.5&	89.2&	94.7&	92.5&	84.4 \\
&SVFit&\textbf{0.018M}     & \textbf{64.8} &	\textbf{92.4}& 	78.0& 	90.0& 	94.4& 	90.8& 	85.1\\
\hline
\multirow{4}{*}{\rotatebox{90}{Large}}&FT & 356M &68 &92.4	&86.6	&90.9	&96.4 &94.7	&88.2 \\
&LoRA & 0.8M& 68.2&	92.3&	85.2	&90.2	&96.2	&94.8	&87.8  \\
&PiSSA & 0.8M& 69.0&	\textbf{92.9}&	85.2	&90.2	&\textbf{96.7}	&\textbf{95.1}	&88.2  \\
&SVFit&\textbf{0.036M }& \textbf{71.4 }&	92.0& 	\textbf{86.3} &	\textbf{90.9} &	96.2 &   94.4  & \textbf{88.5}  \\
\hline
\end{tabular}
\end{table*}

\begin{table*}[t]
\caption{Performance comparison of various fine-tuning methods on the image classification task using ViT-base and ViT-large models across different datasets. Accuracy (\%) is reported after ten epochs. Avg. represents the average accuracy across all datasets. The best performance for each dataset is highlighted in bold}
\label{tab:Image Classification}
\centering
\begin{tabular}{llcccccccccc}
\hline
&\multirow{2}*{Method }& \#  Trainable  & \multirow{2}*{OxfordPets} & \multirow{2}*{CIFAR10} &  \multirow{2}*{DTD} & \multirow{2}*{EuroSAT} &\multirow{2}*{RESISC45}&\multirow{2}*{StanfordCars} & \multirow{2}*{FGVC} &\multirow{2}*{CIFAR100}& \multirow{2}*{Avg.} \\
& ~ &Parameters  & ~  & ~  & ~  & ~  & ~  & ~  & ~ &~ &~  \\
\hline
\multirow{5}{*}{\rotatebox{90}{Base}}&Head & - &90.3 &96.4 &69.8 &88.7 &74.2 &25.8 &17.4 &84.3 &68.4 \\
&FT & 85.8M &93.1 &\textbf{98.9} &77.7 &\textbf{99.1} &\textbf{96.1} &79.8 &\textbf{54.8} &\textbf{92.4} &\textbf{86.5} \\
&LoRA & 0.3M &93.2 &98.8 &75.0 &98.4 &92.7 &45.4 &25.2 &92.0 &77.6 \\
&PiSSA & 0.3M &95.9 &98.6 &78.7 &98.7 &95.5 &67.1 &47.6 &91.2 &84.2 \\
&SVFit & \textbf{0.018M} &\textbf{97.0} &98.8 &\textbf{80.5} &98.6 &93.0 &67.2 &47.9 &91.6 &84.3 \\
\hline
\multirow{5}{*}{\rotatebox{90}{Large}}&Head & - &91.1 &97.8 &73.3 &92.6 &82.0 &37.9 &24.6 &84.3 &73.0 \\
&FT & 303.3M &94.4 &99.2 &81.8 &\textbf{99.0} &\textbf{96.4} &\textbf{88.9} &\textbf{68.3} &93.6 &\textbf{90.2} \\
&LoRA & 0.8M &94.8 &99.1 &81.8 &98.6 &94.7 &73.3 &42.3 &\textbf{94.9} &84.9 \\
&PiSSA & 0.8M &96.7 &98.8 &77.8 &98.8 &95.7 &86.7 &62.6 &92.6 &88.7 \\
&SVFit & \textbf{0.036M} &\textbf{97.8} &\textbf{99.3} &\textbf{83.4} &98.7 &95.2 &83.3 &57.8 &93.9 &88.7 \\
\hline
\end{tabular}
\end{table*}

\begin{table}[t]
\caption{Hyperparameter setup for image classification of SVFit}
\label{tab:hp2}
\centering
\setlength{\tabcolsep}{2pt} 
\begin{tabular}{ll|cccc}
\hline
~ & Hyperparameter & OxfordPets & CIFAR10 & DTD & EuroSAT \\
\hline
\multirow{6}{*}{\rotatebox{90}{Base}} & Rank & \multicolumn{4}{c}{768} \\
 & Epochs & \multicolumn{4}{c}{10} \\
 & LR Scheduler & \multicolumn{4}{c}{Linear} \\
 & Weight Decay & \multicolumn{4}{c}{0.01} \\
 & Trainable Matrices & \multicolumn{4}{c}{$W_{Q},W_{V}$} \\
 & Learning Rate & 0.04 & 0.04 & 0.04 & 0.2 \\
\hline
\multirow{6}{*}{\rotatebox{90}{Large}} & Rank & \multicolumn{4}{c}{768} \\
 & Epochs & \multicolumn{4}{c}{10} \\
 & LR Scheduler & \multicolumn{4}{c}{Linear} \\
 & Weight Decay & \multicolumn{4}{c}{0.01} \\
 & Trainable Matrices & \multicolumn{4}{c}{$W_{Q},W_{V}$} \\
 & Learning Rate & 0.06 & 0.03 & 0.03 & 0.1 \\
\hline
 ~ & Hyperparameter & RESISC45 &StanfordCars & FGVC &CIFAR100  \\
\hline
\multirow{6}{*}{\rotatebox{90}{Base}} & Rank & \multicolumn{4}{c}{768} \\
 & Epochs & \multicolumn{4}{c}{10} \\
 & LR Scheduler & \multicolumn{4}{c}{Linear} \\
 & Weight Decay & \multicolumn{4}{c}{0.01} \\
 & Trainable Matrices & \multicolumn{4}{c}{$W_{Q},W_{V}$} \\
 & Learning Rate & 0.2 & 0.4 & 0.2 & 0.04 \\
\hline
\multirow{6}{*}{\rotatebox{90}{Large}} & Rank & \multicolumn{4}{c}{768} \\
 & Epochs & \multicolumn{4}{c}{10} \\
 & LR Scheduler & \multicolumn{4}{c}{Linear} \\
 & Weight Decay & \multicolumn{4}{c}{0.01} \\
 & Trainable Matrices & \multicolumn{4}{c}{$W_{Q},W_{V}$} \\
 & Learning Rate& 0.07 & 0.4 & 0.3 & 0.1 \\
\hline
\end{tabular}
\end{table}

\subsection{Natural Language Understanding}
\textbf{Models and Datasets.} We evaluate our method on the GLUE benchmark (General Language Understanding Evaluation). This comprehensive natural language understanding assessment encompasses various tasks, including sentence relationship recognition, sentiment analysis, and natural language reasoning~\cite{wang2018glue}. For systematic evaluation, we select six tasks: CoLA~\cite{COLA}, STS-B~\cite{stsb}, RTE~\cite{rte}, MRPC~\cite{MRPC}, SST-2~\cite{sst2}, and QNLI~\cite{QNLI}. We implement SVFit for fine-tuning RoBERTa-base, which has 12 layers with a hidden size of 768, totaling 125 million parameters, and RoBERTa-large, which has 24 layers with a hidden size of 1024, totaling 356 million parameters~\cite{liu2019roberta}.

\textbf{Implementation Details.}  For all six datasets in GLUE, we tune the hyperparameters for learning rates and scaling values. Following the experimental setup used in previous studies~\cite{zi2023delta,lora}, we fine-tune only the query and value weights in each transformer block while fully fine-tuning the classification head. For both models, the rank of LoRA is set to 8, and the rank of SVFit is set to 768. Due to time constraints and budget limitations, we omit the time-intensive MNLI and QQP tasks, thereby forgoing the MNLI trick for MRPC, RTE, and STS-B tasks. Consistent with prior work~\cite{vera,gao2024parameter}, we report the number of trainable parameters in the fine-tuned layers, explicitly excluding the classification head, which is trained in a standard way. The results are averaged over five different random seeds. Additional details are provided in Table \ref{tab:hp1}.

\textbf{Results.} Based on the results presented in Table \ref{tab:GLUE}, our proposed SVFit method demonstrates superior performance in several key areas compared to both traditional fine-tuning (FT) and other PEFT methods such as LoRA, DyLoRA, AdaLoRA, and PiSSA. For RoBERTa-base, SVFit achieves the highest Matthew’s correlation coefficient (MCC) for CoLA and the highest Pearson correlation coefficient (PCC) for STS-B, indicating its effectiveness in classification and regression tasks. While it falls slightly behind in accuracy (ACC) for MRPC and SST-2, it maintains competitive performance across all tasks, resulting in a robust overall average. For RoBERTa-large, SVFit outperforms other methods on CoLA, RTE, and MRPC and achieves comparable performance on other tasks. Its significant improvement in MCC for CoLA is noteworthy, reflecting its strength in complex language understanding tasks. The results suggest that SVFit's approach to fine-tuning, with its unique parameterization and efficient use of trainable parameters, offers a balanced and effective alternative to existing methods. Additionally, the method's reduced number of trainable parameters demonstrates its potential for achieving high performance with lower computational costs.

\subsection{Image Classification}
\textbf{Models and Datasets.} We assess SVFit on the image classification task using both the base and large versions of the widely adopted Vision Transformer (ViT) model~\cite{vit}, pre-trained on the ImageNet-21K dataset~\cite{ridnik2021imagenet}. To ensure a comprehensive evaluation, we employ a diverse set of datasets, including OxfordPets~\cite{OxfordPets}, CIFAR10~\cite{CIFAR10}, DTD~\cite{DTD}, EuroSAT~\cite{Eurosat}, RESISC45~\cite{RESISC45}, StanfordCars~\cite{StandfordCars}, FGVC~\cite{FGVC}, and CIFAR100~\cite{CIFAR10}.

\textbf{Implementation Details.} We evaluated the performance of LoRA, PiSSA, and SVFit applied to the query and value layers of the ViT, in addition to two baseline approaches: full fine-tuning (FT) and training only the classification head (referred to as Head). Consistent with our GLUE benchmark setup, the rank of LoRA is set to 8 and the rank of SVFit to 768 for both models. Learning rates were meticulously tuned for all methods, with the maximum training epoch limited to 10. The reported parameter counts exclude the classification head, which is trained across all methods. Further details are provided in  Table \ref{tab:hp2}.

\begin{figure*}[htbp]
	\centering
   \includegraphics[width=6in]{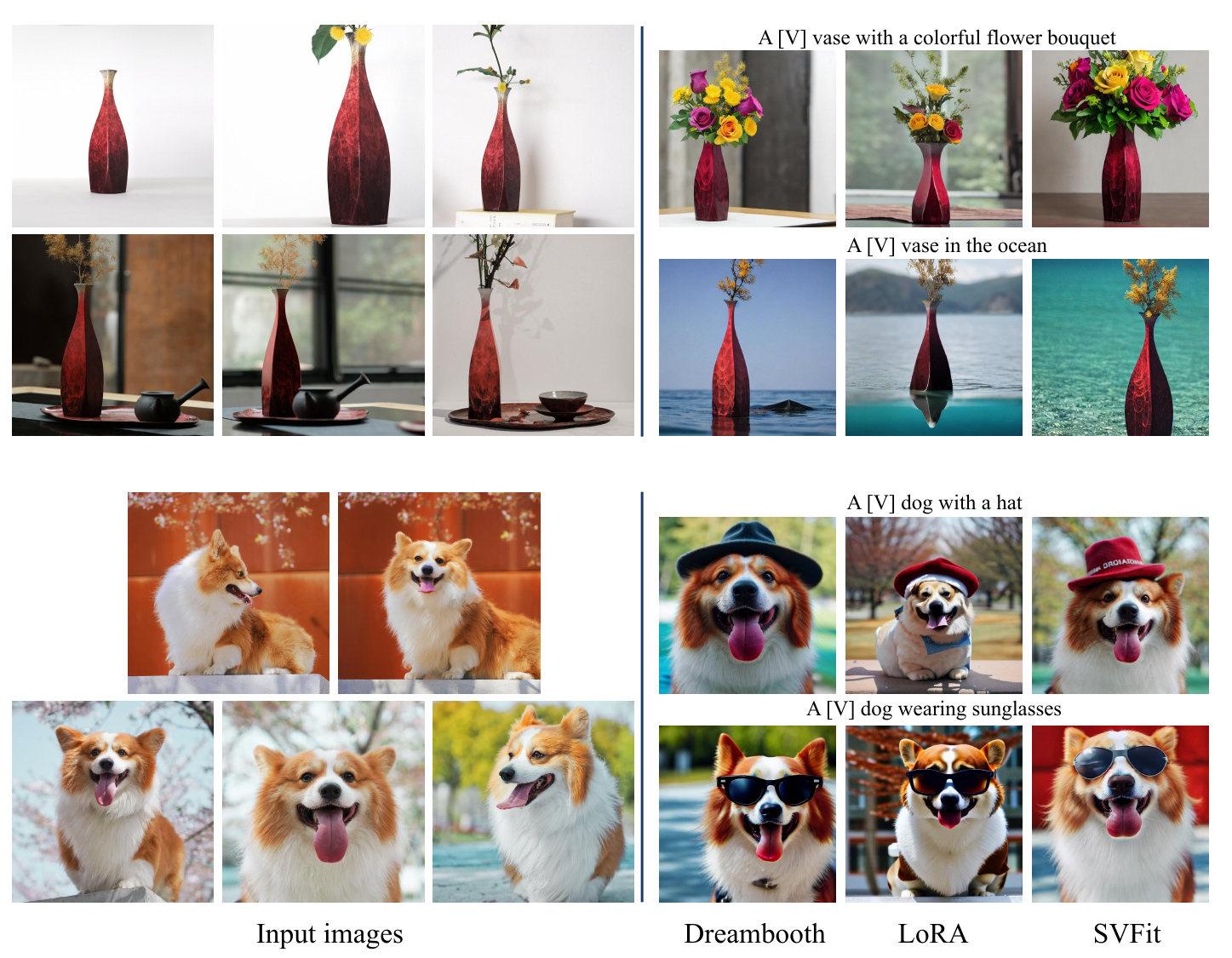}
	\caption{Randomly selected samples from DreamBooth, LoRA, and SVFit for the subject-driven generation task.}
		\label{fig:ti}
\end{figure*}

\textbf{Results.} Table \ref{tab:Image Classification} illustrates the performance of various fine-tuning methods on image classification tasks using ViT-base and ViT-large models across various datasets. Notably, our proposed SVFit method demonstrates strong performance with both model sizes. For the ViT-base, SVFit achieves the highest accuracy on OxfordPets (97.0\%) and DTD (80.5\%) and performs competitively on other datasets, resulting in an overall average accuracy of 84.3\%. While FT achieves the highest average accuracy (86.5\%), it requires significantly more trainable parameters (85.8M) compared to SVFit's minimal 0.018M parameters. For ViT-large, SVFit again stands out by achieving the highest accuracy on OxfordPets (97.8\%), CIFAR10 (99.3\%), DTD (83.4\%), and performs well across other datasets with an average accuracy of 88.7\%. FT still yields the highest average accuracy (90.2\%) but at the cost of training 303.3M parameters, whereas SVFit uses only 0.036M parameters. These results underscore the efficiency and effectiveness of SVFit, particularly in scenarios where computational resources and parameter efficiency are critical. By maintaining competitive performance while dramatically reducing the number of trainable parameters, SVFit offers a compelling alternative to traditional fine-tuning methods.

\subsection{Dreambooth}
\textbf{Models and Datasets.} Following~\cite{ruiz2023dreambooth}, we evaluate our method on the subject-driven text-to-image generation task. By fine-tuning Stable Diffusion v1.5~\cite{SD} using DreamBooth, we can generate diverse images of a subject instance in various environments, maintaining high preservation of subject details and realistic interactions between the scene and the subject. We compare our approach with LoRA and DreamBooth~\cite{ruiz2023dreambooth}, ensuring fairness by randomly selecting generated images from both methods. For fine-tuning, we use the dataset introduced in DreamBooth~\cite{ruiz2023dreambooth}, which includes five or six images per subject for training.

\textbf{Implementation Details.} Both LoRA and our method use the same loss function as in DreamBooth. For DreamBooth and LoRA, we apply the best hyperparameter setup in the original paper~\cite{ruiz2023dreambooth}.

\textbf{Results.} In Fig. \ref{fig:ti}, we present a comparative analysis of image generation using Dreambooth, LoRA, and SVFit across multiple scenarios. SVFit consistently demonstrates superior subject detail preservation and context integration. For instance, in the "vase with a colorful flower bouquet" scenario, SVFit accurately retains the intricate details of the flowers and vase while seamlessly blending them with the background. In contrast, the results from Dreambooth and LoRA exhibit noticeable artifacts and inconsistencies in subject representation and background interaction. Overall, the visual comparison clearly illustrates SVFit's enhanced capability in generating high-fidelity images that closely follow the given prompts, effectively preserving subject details and ensuring realistic environmental interactions.

 \begin{figure*}[t]
	\centering
  \includegraphics[width=7in]{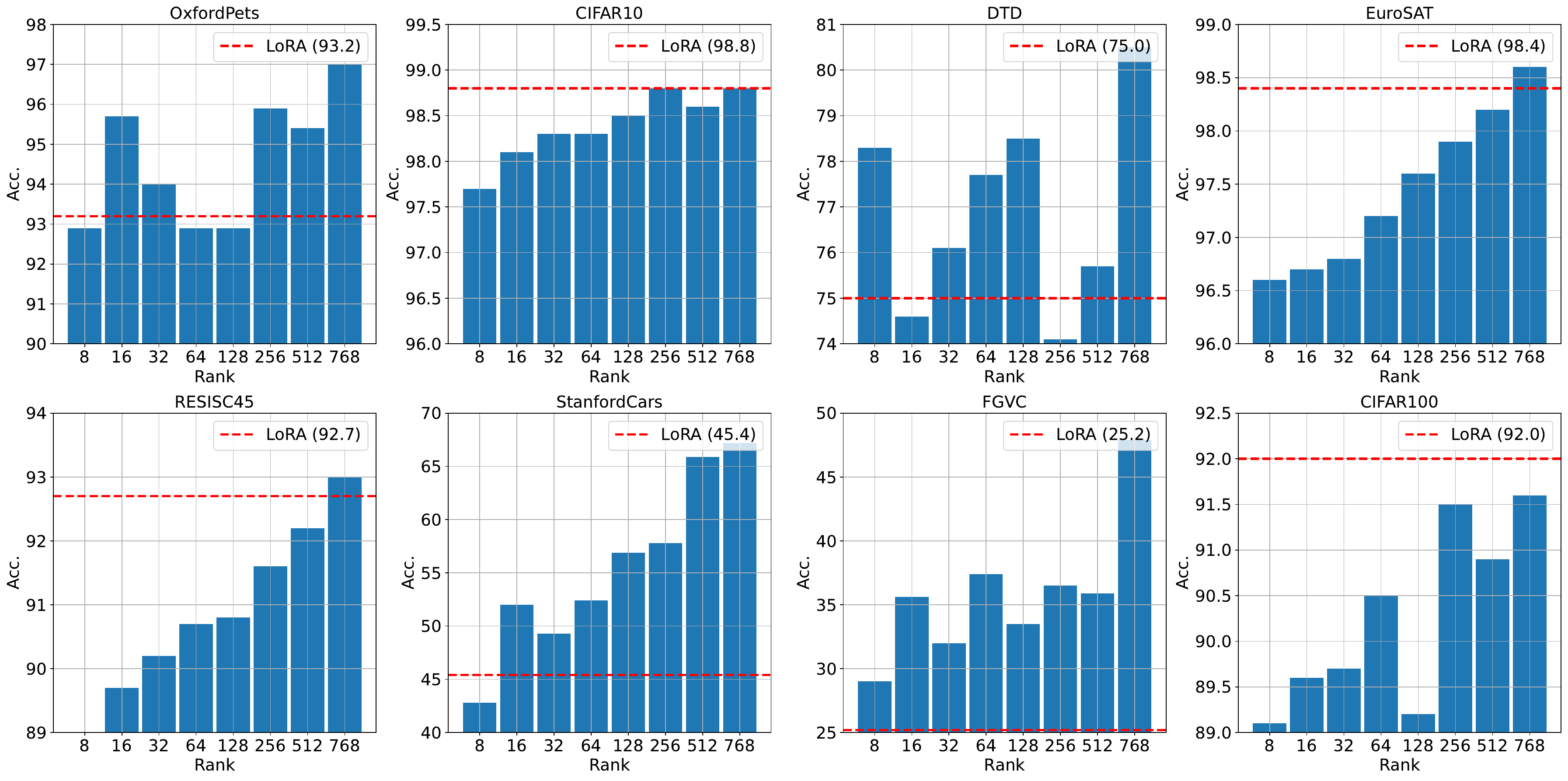}
	\caption{Performance of SVFit fine-tuning for ViT-base model on image classification tasks across different parameter budget levels. The $x$-axis represents the rank, and the $y$-axis is the evaluation index of different datasets.}
		\label{fig:2}
\end{figure*}

 \begin{figure*}[t]
	\centering
  \includegraphics[width=7in]{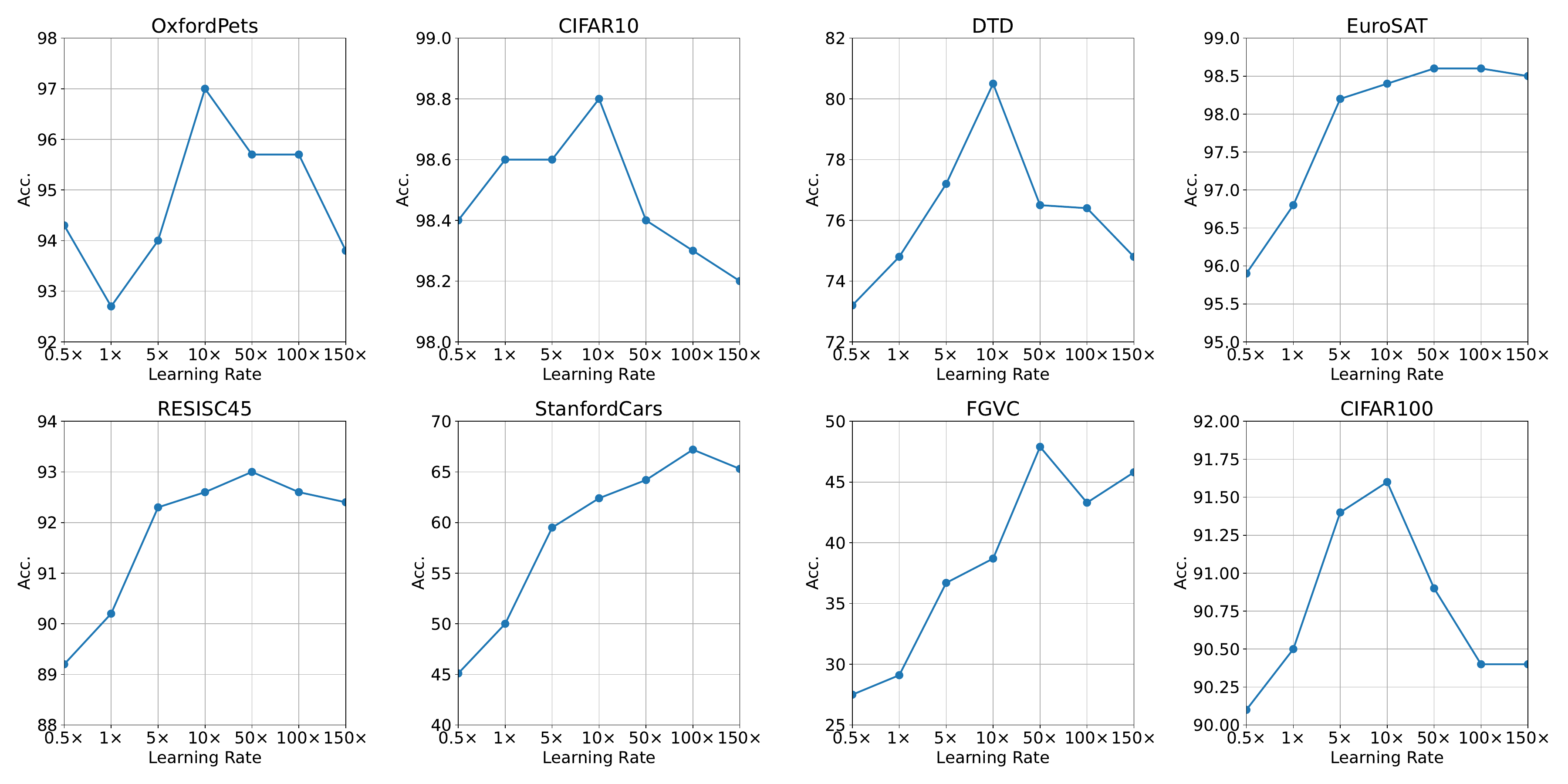}
	\caption{Performance of SVFit fine-tuning for ViT-base model on image classification tasks across different learning rate. The $x$-axis represents the learning rate, and the $y$-axis is the evaluation index of different datasets.}
		\label{fig:3}
\end{figure*}

\subsection{Different Budget Levels}
We analyzed the performance of SVFit fine-tuning on the ViT-base model for image classification tasks under different parameter budget levels. The specific results are shown in Fig. \ref{fig:2}. We employed various ranks \( r = \{8, 16, 32, 64, 128, 256, 512, 768\} \), corresponding to 0.2K, 0.4K, 0.8K, 1.5K, 3.1K, 6.1K, 12.3K, and 18.4K trainable parameters, respectively. For LoRA, we used a baseline rank of \( r = 8 \), corresponding to 294.9K trainable parameters. The experimental results demonstrate that our method effectively balances the number of trainable parameters and accuracy. For instance, on the OxfordPets dataset, our method achieves an accuracy of approximately 93.2\% at the lowest rank of 8 while maintaining or improving accuracy with higher ranks. Similarly, for the CIFAR10 dataset, our method achieves an accuracy of around 98.8\% at rank 8, with further gains observed as the rank increases. This trend is consistent across other datasets. Our method significantly improves over the baseline across various datasets and rank levels, maintaining high performance with fewer trainable parameters.

\subsection{Analysis of Learning Rate}
Adjusting the learning rate is a crucial step in fine-tuning.   Our method requires a larger learning rate than LoRA since our initialization strategy has already initialized most of the model’s parameters to a certain extent, and a larger learning rate can help quickly adapt the remaining parameters to the new tasks.   We perform a learning rate search using our method, as shown in Fig. \ref{fig:3}. A learning rate 10× greater than pre-training yields the best results across multiple datasets.   For instance, on the OxfordPets dataset, the accuracy peaks at approximately 93.2\% with a 10× learning rate, while both lower (0.5×, 1×, 5×) and higher learning rates (50×, 100×, 150×) result in decreased performance.   However, for EuroSAT, an accuracy of around 98.6\% is achieved at 10×, with lower and higher rates underperforming. These findings demonstrate that while a substantial increase in the learning rate from pre-training is generally beneficial, the optimal rate can vary significantly depending on the specific dataset.

\section{Conclusion}
In this work, we introduced SVFit, a novel PEFT method that enhances initialization by leveraging SVD to initialize low-rank matrices derived from pre-trained weights. SVFit focuses on training only the most significant top-\(r\) singular values, significantly reducing the number of trainable parameters while ensuring efficient fine-tuning and preserving the model's core capabilities. Our theoretical analysis demonstrates how this approach enables rapid adaptation by effectively capturing essential information from pre-trained models and efficiently learning new domain-specific knowledge with minimal parameters. SVFit has been evaluated across various tasks, including natural language understanding, image classification, and subject-driven text-to-image generation, consistently outperforming LoRA and other recent state-of-the-art techniques such as PiSSA in both efficiency and effectiveness. Future work will focus on extending SVFit to more complex tasks, optimizing singular value selection, and further enhancing performance through dynamic parameter budget allocation to broaden its applicability across diverse domains.

\section*{Acknowledgements}
This research was partially funded by the National Natural Science Foundation of China under grants 62220106008, 62306067, and U20B2063, as well as the Sichuan Science and Technology Program under grant 2024NSFSC1463. Additional support was provided by the Sichuan Province Innovative Talent Funding Project for Postdoctoral Fellows (Project BX202311) and the China Postdoctoral Science Foundation (Project 2022M720660).

\bibliographystyle{IEEEtran}
\bibliography{bbb}

\end{document}